\documentclass{article}

\usepackage{arxiv}

\usepackage{tabularx}

\usepackage[utf8]{inputenc} 
\usepackage[T1]{fontenc}    
\usepackage{hyperref}       
\usepackage{url}            
\usepackage{booktabs}       
\usepackage{amsfonts}       
\usepackage{nicefrac}       
\usepackage{microtype}      
\usepackage{lipsum}
\usepackage{graphicx}
\graphicspath{ {./images/} }
\usepackage{hyperref}
\usepackage[export]{adjustbox}
\usepackage{float}
\usepackage{geometry}
\usepackage{booktabs}
\usepackage{caption}
\usepackage{setspace} 
\usepackage{etoolbox}
\usepackage{float} 
\usepackage{multirow}

\title{Large Language Models in Analyzing Crash Narratives - A Comparative Study of ChatGPT, BARD and GPT-4 
}

\author{
 Maroa Mumtarin \\
  Department of Civil, Construction, and Environmental Engineering\\
  Iowa State University\\
  Ames, IA, 50011 \\
  \texttt{maroa@iastate.edu} \\
   \And
 Md Samiullah Chowdhury \\
  Department of Civil and Environmental Engineering\\
  South Dakota State University \\
  Brookings, SD, 57007 \\
  \texttt{mdsamiullah.chowdhury@jacks.sdstate.edu} \\
  \And
 Jonathan Wood \\ Assistant Professor\\
  Department of Civil, Construction, and Environmental Engineering\\
  Iowa State University\\
  Ames, IA, 50011 \\
  \texttt{jwood2@iastate.edu} \\
}

\usepackage{titlesec} 
\titlespacing{\section}{0pt}{3pt}{3pt} 
\titlespacing{\subsection}{0pt}{2pt}{2pt} 
\begin{document}
\maketitle
\vspace{3.5em}
\begin{abstract}
In traffic safety research, extracting information from crash narratives using text analysis is a common practice. With recent advancements of large language models (LLM), it would be useful to know how the popular LLM interfaces perform in classifying or extracting information from crash narratives. To explore this, our study has used the three most popular publicly available LLM interfaces- ChatGPT, BARD and GPT4. This study investigated their usefulness and boundaries in extracting information and answering queries related to accidents from 100 crash narratives from Iowa and Kansas. During the investigation, their capabilities and limitations were assessed and their responses to the queries were compared. Five questions were asked related to the narratives - 1) Who is at-fault? 2) What is the manner of collision? 3) Has the crash occurred in a work-zone? 4) Did the crash involve pedestrians? and 5) What are the sequence of harmful events in the crash? For questions 1 through 4, the overall similarity among the LLMs were 70\%, 35\%, 96\% and 89\%, respectively. The similarities were higher while answering direct questions requiring binary responses and significantly lower for complex questions. To compare the responses to question 5, network diagram and centrality measures were analyzed. The network diagram from the three LLMs were not always similar although they sometimes have the same influencing events with high in-degree, out-degree and betweenness centrality. This study suggests using multiple models to extract viable information from narratives. Also, caution must be practiced while using these interfaces to obtain crucial safety related information. 
\end{abstract}


\section{Introduction}
Motor vehicle traffic (MVT) incidents are a major public health concern in the United States. They result in millions of injuries and thousands of deaths each year. Concerningly, the number of MVT fatalities has been increasing, with a 10.5\% increase from the previous year in 2021 \cite{nhtsa2021traffic}. To document traffic incidents and enhance road safety, state-level and local police agencies in all 50 states collect data and publish crash reports on MVT incidents. While the state agencies collect data on mainly the interstates and highways, local agencies are responsible for collecting data on crashes within their authority \cite{lopez2022police}. These crash reports include a text-based 'crash narrative' written by the investigating officer at the scene of the incident. With the recent advancements of large language models (LLM) and their extensive capability in text analysis, new tools are available for researchers who are trying to implement Machine Learning (ML) and Natural Language Processing (NLP) techniques to address various road safety issues. Although ML models have been used for years to analyze and extract information from crash narratives, training an ML model demands a vast amount of data, and the accuracy is contingent on this data amount and quality. In contrast, LLM models are pre-trained on a vast array of knowledge and fine-tuned for understanding narrations\cite{ouyang2022training}. It could immensely aid the agencies lacking large datasets for training the ML models. Also, a substantial amount of computational cost is associated with training ML models adequately. Furthermore, each type of work, either text analysis or prediction, might require different models for investigating diverse research questions. Publicly available LLM interfaces like ChatGPT, BARD and GPT-4 can be of help in these cases. Besides, current ML models often struggle to retrieve information directly for complex questions- such as creating a sequence of events from the crash narrative to understand the overall scenario of the crash. With the new release of LLM models, it is essential to explore their capabilities and limits in extracting critical information from crash narratives that could be used in updating crash databases. Knowing their boundaries in generating resources would be helpful for future research and developments. At the same time, it would be beneficial to compare the similarities, dissimilarities, and overall usability of different LLMs in transportation safety research. Hence, this study has three main objectives: 1. Extract information to update crash databases from crash narratives by using LLM interfaces, 2. Assess the capabilities and limitations of the LLM interfaces in answering questions from crash narratives, 3. Explore similarities and dissimilarities in the most common LLM interfaces in analyzing crash narratives. Three most common LLM interfaces: ChatGPT, BARD and GPT-4 is used in this study to compare the responses. This study will help to understand the effectiveness and limitations of LLMs in analyzing crash narratives to extract significant information. It is worth mentioning that some LLM interfaces explored in this study are being constantly improved for safer responses from them. So, this study will guide the path to exploring the capabilities of LLMs as they improve.

\section{Literature Overview}
Crash narratives are often an effective source of accident analysis as they provide valuable information leading to crashes. These include circumstances and contributing factors leading to a crash, such as the weather conditions, the location of the crash, the number of vehicles involved, the injuries sustained, the actions of the drivers involved, the DUI state of the drivers involved and the presence of any road hazards\cite{montella2013crash}.

In recent years, researchers have used machine learning and natural language processing (NLP) techniques to extract information from crash narratives. These techniques can identify specific terms and phrases in the narratives, which can then be used to classify crashes according to their types, making safety analysis more efficient and effective. For example, one study used logistic regression to identify speeding-related crashes from crash narratives. The study found that logistic regression could accurately identify speeding-related crashes with an accuracy of 89.7\% \cite{fitzpatrick2017investigation}. Another study used machine learning techniques to identify pedestrian-related crashes from the narratives \cite{das2020application}. A recent study has developed a framework for the automatic classification of road accidents using a combination of NLP and ML. This study found that the most accurate and explainable solution was to combine Hierarchical Dirichlet Process (HDP) topic modeling and random forest classification\cite{valcamonico2022framework}. Several other studies have also used crash narratives in conjecture with text mining, ML and NLP techniques to study specialty crashes such as railroad, motorcycle, and work-zone related crashes \cite{brown2015text,das2021topic,sayed2021identification}.These classification techniques help isolate crashes according to their types, thus making the safety analysis more efficient. However, ML and NLP techniques have drawbacks due to their fundamental reliance on word-based methods. With no regard for the semantics of text, they cannot understand the text's contextual meaning as humans naturally do. However, their improvements can be facilitated through an access to vast amounts of knowledge about the real world and the specific domain of discourse \cite{cambria2014jumping}. Recent developments in AI (Artificial Intelligence) and large language models such as ChatGPT, BARD and GPT-4 can be utilized in this scenario as they are pre-trained with vast information. The ability of these AIs to classify the text and extract latent information can be used to classify crashes and discern the contributing factors from crash narratives. This information can subsequently be used to develop targeted road safety interventions \cite{zheng2023chatgpt}. Although AI is predicted to be of major use in a vast array of transportation domains like autonomous driving, human-vehicle interaction, and intelligent transportation systems \cite{du2023chat}, this technology is relatively new and publications using AI in transportation safety are still in its growing stage. Recently, however, a notable study applied Bidirectional Encoder Representations from Transformers (BERT – an LLM developed by Google AI in 2018) to classify traffic injury types. The study used over 750,000 unique crash narrative reports and found that BERT was able to classify traffic injury types with an accuracy of 84.2\%.\cite{oliaee2023using} We further extend these efforts and assess the capabilities and limitations of the latest publicly available LLMs in the domain of crash analysis using only the narratives from publicly available crash reports.

The three LLM interfaces chosen for experimentation in this study are pre-trained generative transformers. Transformer is a novel neural network (NN) architecture based on a self-attention mechanism that is especially well suited for language understanding. It has been shown in a study that the NN architecture outperforms both recurrent and convolutional models on translational tasks \cite{vaswani2017attention} The most popular LLM interface deployed in this study is- ChatGPT, introduced by OpenAI in November 2022 as a modified model for conversation in the GPT-3.5 series. The GPT-3.5 series is an accumulation of models that were trained on texts and codes until September 2021 using approximately 175B parameters \cite{wu2023brief}. Concurrently, GPT-4 was also developed by OpenAI, which is a significant multimodal model that not only outperforms the GPT-3.5 models in analytical capabilities and precision but is also optimized for conversational inputs. In contrast, Google's experimental conversational AI, BARD (Bidirectional Encoder Representations from Transformers), is based on LaMDA: Language Models for Dialog Applications. LaMDA, a family of Transformer-based neural language models dedicated to conversation. It has been trained on 1.56 trillion words from public dialogue data and web text, comprising approximately 137 billion parameters\cite{thoppilan2022lamda}. In terms of token management, ChatGPT, BARD and GPT-4 is able to process up to 4096, 4096 and 8192 tokens respectively. A key distinction between ChatGPT, BARD, and GPT-4 lies in their knowledge cutoff. While the information within ChatGPT and GPT-4 is current up to September 2021, BARD's knowledge is more recent, as it has a continuous updating structure.

Although these LLMs present the prospect of revolutionizing research, they have some potential risks. ChatGPT has been analyzed for such risks in research, and it has been found to provide incorrect information and make improper assumptions at times \cite{oviedo2023risks}. These results from insufficient training of the LLMs, which is going to improve with time. As with the GPT-4 (successor of ChatGPT), the OpenAI claims to have done significant improvements in training for providing safer and more accurate results. But for the time being, while we implement LLMs in our research, we must exercise caution and careful consideration. Hence to ensure the reliability and accuracy of our research work, we have provided LLMs with explicit guidance for responding to our queries. Thus, we effectively minimized the possibility of generating irrelevant or potentially misleading information.

\section{Data Description}
Publicly available crash data from the websites of two highway agencies- Iowa State Patrol and Kansas Highway Patrol were used in this study. The Iowa State Patrol website (\href{http://www.overleaf.com}{https://accidentreports.iowa.gov/}) holds minimal crash reports of all incidents under its jurisdiction from all over Iowa. It holds these reports for up to 15 days of that event with a narrative of the incident. Similarly, the Kansas Highway Patrol website (\href{http://www.overleaf.com}{https://www.kansas.gov/khp-crashlogs/search.do}) records crash logs from all over Kansas state, except the turnpike county. Their records department maintains the crash report on the website for 14 days from the incident happening, and the report has a ‘crash narrative’ field. 

As the primary objective of this study was to analyze only the crash narrative, full crash reports were not requested from these authorities. A total of 100 crash narratives were used in this study, dated from 20 April 2023 to 14 May 2023. 15 narratives were from the Iowa State Patrol, and 85 narratives were from Kansas Highway Patrol. 

The responses from the LLM interfaces were collected between the months of May-2023 and June-2023. This data collection process did not use any Application Programming Interface (API); rather, web-based chat interfaces were used to collect the feedback.

\section{Method}
Our research approach involved collecting and analyzing one hundred publicly available crash narratives and then using the LLM interfaces to run five distinct queries pertaining to traffic accidents. It is worth mentioning that, while answering queries, LLM interfaces provide redundant information, potentially convoluting the process of getting the answers to align with the police-reported crash database. To circumvent this issue, we have incorporated a list of categories in the prompt for the LLM to answer within. These categories are taken from the Iowa DOT's (Department of Transportation) official crash reporting guide \cite{iowadot_2023_guide}. We have asked several types of questions to the LLMs and documented the answers. The rationale for asking these specific questions are provided in the subsequent sections. 

The first question is- 
\begin{itemize}
\item Who is at fault in this accident? [crash narrative]
\end{itemize}
Whenever crashes happen, the police document necessary details of the crash in the crash report for further investigation of the accident. This information is particularly important for law enforcement and insurance disputes. In safety research, this at-fault information is also important because it helps identify the contributing factors behind the crash. Written descriptions or crash narratives by the investigating officer are critical in determining this at-fault assignment. Hence, we have checked whether these LLMs can provide information regarding at-fault assignments.

The second question is- 
\begin{itemize}
\item What is the manner of crash collision in this accident? [crash narrative]
\end{itemize}
The manner of a crash collision indicates how the first harmful event of a crash took place. This field is important for measuring occupant injury and structural defects \cite{mmucc_guideline}. Moreover, this field is important for preparing collision diagrams to identify similar accident patterns at any particular location. Typically, most police-reported crash narratives contain sufficient information about the circumstances leading to crash occurrence, so the manner of crash collision is easily derivable from the narratives. To facilitate this extraction, we have provided a list of categories for the manner of crash collision per the Iowa accident reporting guide \cite{iowadot_2023_guide}. Then the LLM was given a prompt to classify the crash narrative according to the list. We have provided around 20-25 narratives at a time and asked the LLM to provide the results in a tabular format. Figure-1 shows the prompt provided to the LLM interfaces for question 2. 
\begin{figure} 
    \centering
    \includegraphics[width= \linewidth]{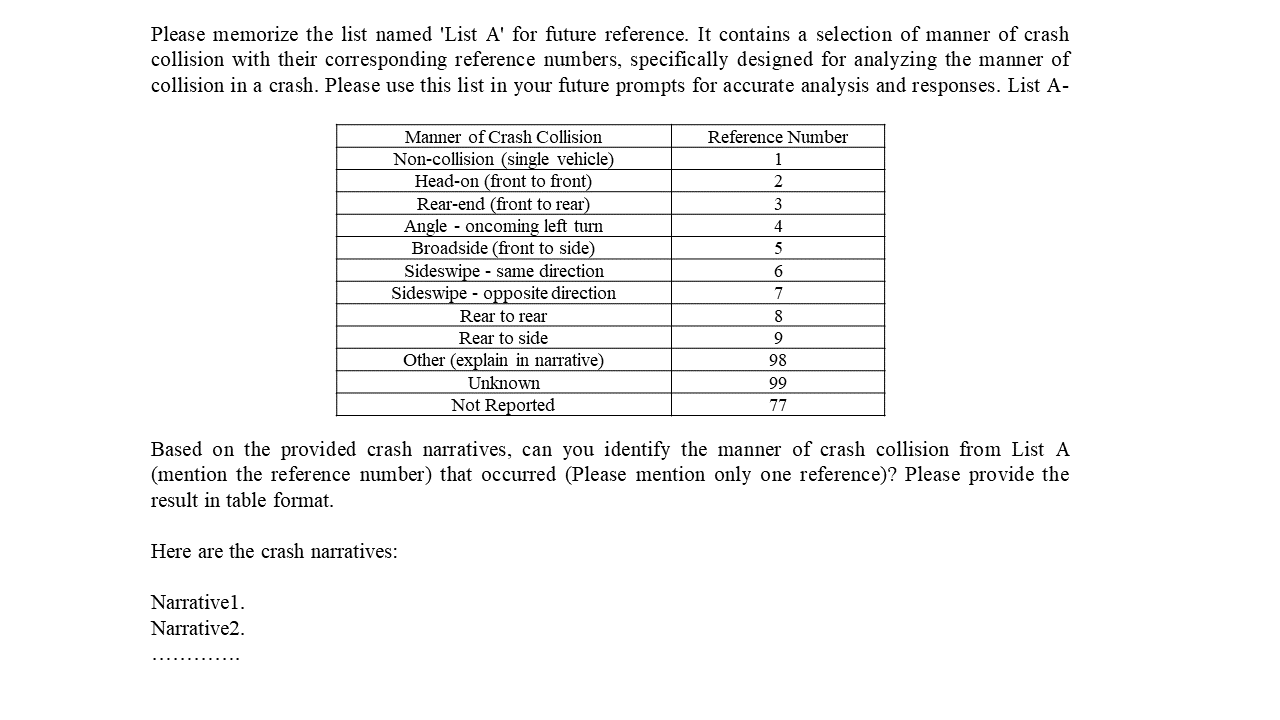}
    \caption{Prompt for Question 2 determining the manner of crash collision}
\end{figure}

In question 3, it was asked whether the crash happened in a work-zone area or not. Work-zone crashes are usually under-represented in crash databases and sometimes unnoticed by the reporting personnel as they are not considered as contributing factors by many law enforcement personnel \cite{sayed2021identification}. Hence, we explored whether this information can be extracted from the narratives.

The prompt for question 3 is-
\begin{itemize}
\item Based on the provided crash narratives, can you identify if the crash has occurred in work-zone or not? Please provide the result in tabular format.
\end{itemize}
In question 4, it was asked if the crash involved any pedestrians. Pedestrians are considered vulnerable road users, and identifying the crashes involving pedestrians facilitates further analysis of the crash causation pattern for pedestrian crashes \cite{yue2020depth}. Pedestrian identification from the narrative would help further to reduce misclassification of crash reports.

The prompt for question 4 is-
\begin{itemize}
\item Based on the provided crash narratives, can you identify if the crash involves a pedestrian? Please provide the result in a tabular format.
\end{itemize}
In question 5, we attempted to give more complex questions to the LLMs to understand their applicability in answering those. So, a prompt was provided to the LLMs to identify the sequence of harmful events in the crash from crash narratives. Traditional ML models will struggle to answer these types of complex questions, where multiple responses are required and event sequence is needed. According to the Model Minimum Uniform Crash Criteria guide, the sequence of events can be collected from the crash narrative. These fields facilitate the identification of the first and most harmful event in a crash. The prompt for question 5 is presented in figure -2. A full list of the sequence of events can be found in Iowa DOT's crash reporting guide \cite{iowadot_2023_guide}. 

\begin{figure} [h] 
    \centering
    \includegraphics[width=\linewidth]{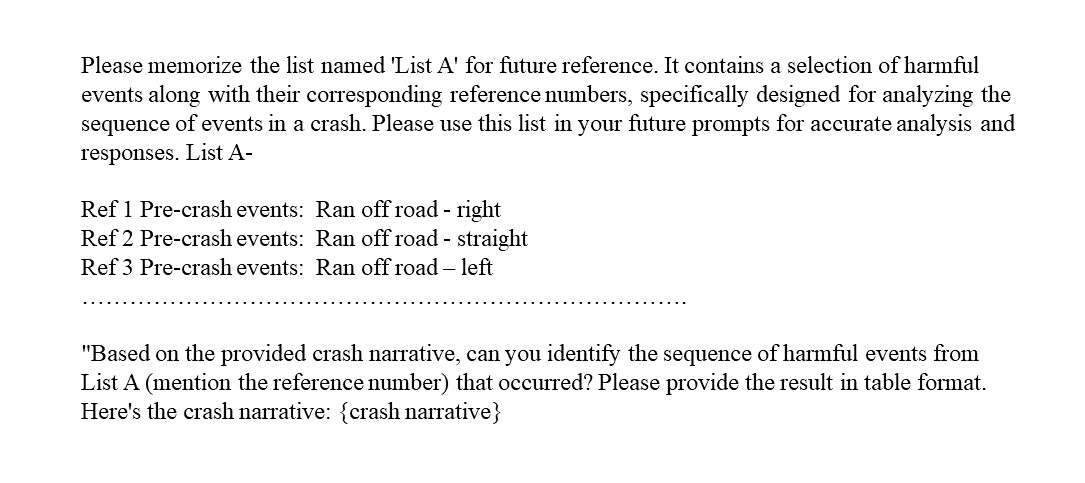}

    \caption{Prompt for Question 5 determining the sequence of events.}
\end{figure}
We have documented the response against these prompts from the three LLM interfaces- ChatGPT, BARD and GPT-4, and then analyzed their responses for a comprehensive study. As the response of the first four queries were either direct responses (such as the manner of collision) or binary answers (yes/no/cannot determine), the comparison was relatively straightforward. For question 5, as it produces 4 to 6 sequences of events for each crash, the direct comparison produced more complexity for analysis. 

To address this, we established a network for the sequence of events produced by each crash for all the LLM interfaces and then used it in network topology to better apprehend the sequences. Previous research has demonstrated the use of network topology in understanding the association of different topics in different types of fatal crashes (i.e., single vehicle/ pedestrian involved) \cite{kwayu2021discovering}. Another study has compared networks of two sets of crash narrative, one involving autonomous vehicle crash with vulnerable road users (VRU), and the other concentrating on autonomous vehicle crash excluding VRUs \cite{kutela2022mining}. 

A network of events can help understand the overall crash scenario in a particular location. For example, if a roadway segment or intersection is selected for diagnostics, creating a network of events for all the crashes in that location can help identify the common pattern of crashes. Furthermore, centrality measures of a network can aid in identifying the issues requiring higher importance for any safety intervention. In our study, we have implemented three types of centrality measures for identifying the influential elements of a network. Degree centrality measures the number of total links in a node and in-degree centrality and out-degree centrality are two properties of a directed network. In-degree centrality measures the number of links directed toward a node, and out-degree centrality measures the number of links generated from a node to connect with others \cite{zhang2017degree}. In this perspective, the nodes with the higher in-degree centrality correspond to common phenomena in crash events. So, the other nodes feeding the higher in-degree centrality nodes should be studied for further safety interventions as those nodes contribute the most in a crash event. On the other hand, out-degree centrality directly identifies the common initiating and secondary events in crashes. For example, if network diagram from multiple crashes from one location shows –‘Pre-crash event: crossed centerline’ has the most out-degree centrality, it should be investigated further to check why this event is more prominent in that location. 

We have also calculated the betweenness centrality. Betweenness centrality is a measure of nodes showing the influence of that node in the overall network. It is calculated by taking the fraction of the shortest path in the network which passes thru that node \cite{newman2005measure}. Betweenness centrality will help to identify the sequences which control the flow of the network.

\section{Results and Discussion}
\label{sec:headings}
\subsection{Query 1}
We have analyzed the zero-shot responses from the LLM interfaces for the 100 crash narratives and compared the results.  The first question answers who is at fault in that accident. Generally, crash narratives are written in such a way that ‘vehicle 1’ tends to be at fault most of the time. Answers from BARD and GPT-4 support this statement most. It is also seen that ChatGPT and BARD have answered ‘cannot determine’ at a higher rate than GPT-4. Table-1 shows the counts of major categories of answers against question-1 for 100 crash narratives, when any two of the interfaces provided the same answers. While creating charts, we have combined some categories where their semantic meaning is same. For example, ‘unit-1’, ‘vehicle 1 driver 1’, ‘vehicle 1’, ‘unit \#1’, ‘vehicle \#1’ are combined to ‘vehicle 1’ category. (Table-1). For question-1 the similarity is 79\% for BARD and GPT-4, 78\% for ChatGPT and BARD, 70\% for GPT4 and ChatGPT. 

In a particularly complex scenario stated below, ChatGPT differed with BARD and GPT-4. In the incident, ChatGPT holds vehicle 3 accountable for the accident but the other two have assigned vehicle 1 as the vehicle at-fault. The street names are redacted in this paper for narrative descriptions.

\textit{'Vehicle 3 was westbound on Roadway X. Vehicle 3 came to a stop in the westbound lane. To try and avoid Vehicle 3, Vehicle 1 drove left of center into the eastbound lane. Vehicle 1 continued across the eastbound lane onto the south shoulder. Vehicle 2 was eastbound on Roadway X. Vehicle 2 attempted to avoid Vehicle 1 by driving onto the south shoulder. Vehicle 1 collided head on with Vehicle 2 on the south shoulder. Both Vehicle 1 and Vehicle 2 came to rest on the south shoulder on Roadway X.'}

In another narrative stated below, BARD assigned vehicle 2 as the vehicle at-fault but the other two  assigned vehicle 1 as at-fault. 

\textit{‘Vehicle 1 and Vehicle 2 were Westbound on Roadway X. Vehicle 1 was catching up to Vehicle 2 and tried to pass Vehicle 2. Vehicle 2 began to make a left turn. Vehicle 1 struck Vehicle 2 on the left side of the truck. Both vehicles traveled South off the highway and came to rest in the South ditch in a windbreak.’}

In all the narratives, where GPT-4 assigned vehicle 1 as at-fault, the other two LLMs agreed.  

\begin{table} [H]
    \centering
    \captionsetup{{skip=5pt}}
    \caption{Count of Similar Answers between Two Interfaces (Question-1)}
    \begin{tabular}{lccc}
        \toprule
        \textbf{Categories} & \textbf{ChatGPT-BARD} & \textbf{BARD-GPT4} & \textbf{GPT4-ChatGPT} \\
        \midrule
        cannot determine & 11 & 5 & 4 \\
        vehicle 1 & 63 & 70 & 64 \\
        vehicle 2 & 2 & 1 & 2 \\
        CMV & 1 & 1 & 0 \\
        Pilot & 1 & 0 & 0 \\
        driver of the vehicle & 0 & 0 & 0 \\
        vehicle 3 & 0 & 1 & 0 \\
        Vehicle & 0 & 0 & 0 \\
        driver of the motorcycle & 0 & 1 & 0 \\
        \midrule
        \textbf{Total} & \textbf{78} & \textbf{79} & \textbf{70} \\
        \bottomrule
    \end{tabular}
\end{table}

\subsection{Query 2}
In question 2, it was asked what is the manner of collision in that narrative and the probable categories were also given to the LLM interfaces. Table-2 shows the results for question-2. 

It is seen that ChatGPT and GPT-4 has the most similarity in answering this question. It could be result of having the same architecture and similarities in the training process of these two LLM models. It is also seen that Bard has elected ‘Sideswipe - same direction’ as the manner of crash in most events, whereas ChatGPT and GPT-4 has chosen ‘Non-collision (single vehicle)’ for most of the answers.

\begin{table} [H]
    \centering
    \caption{Similarity Count among LLM interfaces for Question 2}
    \resizebox{\textwidth}{!}{%
    \begin{tabular}{p{3.5cm}ccc|ccc}
        \toprule
        \textbf{Answer Categories} & \multicolumn{3}{c}{\textbf{Count of Answers}} & \multicolumn{3}{c}{\textbf{Count of Similar Answers between Two Interfaces (Question-2)}} \\
        \cmidrule(lr){2-4}
        \cmidrule(lr){5-7}
        & \textbf{ChatGPT} & \textbf{Bard} & \textbf{GPT-4} & \textbf{ChatGPT-BARD} & \textbf{BARD-GPT4} & \textbf{GPT4-ChatGPT} \\
        \midrule
        Non-collision (single vehicle) & 37 & 17 & 46 & 12 & 17 & 36 \\
        Head-on (front to front) & 16 & 10 & 10 & 8 & 8 & 8 \\
        Rear-end (front to rear) & 25 & 16 & 17 & 15 & 15 & 16 \\
        Angle - oncoming left turn & 5 & 14 & 2 & 4 & 2 & 0 \\
        Broadside (front to side) & 5 & 0 & 10 & 0 & 0 & 2 \\
        Sideswipe - same direction & 5 & 22 & 8 & 1 & 1 & 3 \\
        Sideswipe – opposite direction & 3 & 3 & 0 & 1 & 0 & 0 \\
        Rear to rear & 0 & 3 & 0 & 0 & 0 & 0 \\
        Rear to side & 1 & 1 & 1 & 0 & 0 & 0 \\
        Other (explain in narrative) & 1 & 13 & 6 & 0 & 2 & 0 \\
        Unknown & 2 & 0 & 0 & 0 & 0 & 0 \\
        Not Reported & 0 & 1 & 0 & 0 & 0 & 0 \\
        \midrule
        \textbf{Total} & \textbf{100} & \textbf{100} & \textbf{100} & \textbf{41} & \textbf{45} & \textbf{65} \\
        \bottomrule
    \end{tabular}%
    }
\end{table}

\subsection{Query 3 \& Query 4}

In question-3, it was asked if the crash occurred in a work-zone area or not. And in question-4, it was asked if the crash involved any pedestrian or not. Both of these questions were expected to provide binary responses. Table-3 shows the results for question-3 and question-4.

\begin{table}[H]
    \centering
    \caption{Similarity Count among LLM interfaces for Question 3 and Question 4}
    \resizebox{\textwidth}{!}
    {
    \begin{tabular}{p{2.5cm}p{2.5cm}ccccccc}
        \toprule
        \multirow{2}{*}{\textbf{Questions}} & \multirow{2}{*}{\textbf{Categories}} & \multicolumn{3}{c}{\textbf{Count of Answers}} & \multicolumn{3}{c}{\textbf{Count of Similar Answers Between Two Interfaces}} \\
        \cmidrule{3-5}
        \cmidrule{6-8}
        & & \textbf{ChatGPT} & \textbf{BARD} & \textbf{GPT-4} & \textbf{ChatGPT-BARD} & \textbf{BARD-GPT4} & \textbf{GPT4-ChatGPT} \\
        \midrule
        \multirow{3}{*}{Question-3} & Yes & 5 & 5 & 3 & 3 & 3 & 3 \\
        & No & 95 & 95 & 97 & 93 & 95 & 95 \\
        & Total & 100 & 100 & 100 & 96 & 98 & 98 \\
        \midrule
        \multirow{3}{*}{Question-4} & Yes & 4 & 7 & 4 & 2 & 1 & 1 \\
        & No & 96 & 93 & 96 & 91 & 90 & 93 \\
        & Total & 100 & 100 & 100 & 93 & 91 & 94 \\
        \bottomrule
    \end{tabular}
    }
\end{table}

GPT-4 identified only 3 crashes as work-zone crash. Both BARD and ChatGPT gave the similar answers for those 3 narratives. Those narratives are (street and city names are redacted in the description):

\textit{1. Vehicle two was stopped in the construction zone. Vehicle one failed to yield to vehicle two and struck vehicle two in the rear.}

\textit{2. Vehicle 1 was traveling eastbound on Roadway X at milemarker 327 in Y City in the Construction Zone. Vehicle 1 made a left hand turn northbound across eastbound lanes. Vehicle 2 struck Vehicle 1 in drivers side door.}

\textit{3. Vehicle 1 was Southbound on Roadway X at a high rate of speed in the left lane. Roadway X from milepost 6.6 is one lane due to a 50mph construction zone. Vehicle 1 swerved into the closed right lane and struck a cut-out section in the road. Vehicle 1 lost control and struck a second cut out section at milepost 6. Driver 1 was ejected. Vehicle 1 and Driver 1 came to rest in the closed right lane.
}

It is obvious from these three narratives that all these are related to work-zones (construction area). But there are some events where only ChatGPT and Bard classified the crashes as work-zone crashes and GPT-4 did not. ChatGPT picked up two events- the presence of a KDOT (Kansas DOT) barrier wall, and a patrol vehicle with emergency lights on as work-zone. Bard also classified two events- the presence of a crash barrel and a driver removing debris from road as work-zone. So, it can be inferred that GPT-4 was more conservative in assigning work-zone crashes compared to others.

In question-4, the query was targeted towards detecting involvement of any pedestrian in the crash. There is just one narrative where three of the interfaces gave the same answer indicating a pedestrian was involved. That narrative is-

\textit{1. Victim exited his vehicle and entered the roadway and was struck by a second vehicle - accident still under investigation.}

It is obvious from the above-mentioned narrative that the crash involves a non-motorist person. 

The amount of pedestrian involvement is higher in the responses from BARD, as it has considered all the crashes involving deer as pedestrian crash. For ChatGPT, when the narrative includes texts like ‘someone ejected from the vehicle’, it has considered it as a pedestrian crash. Figure-3 shows the crash narratives involving a deer and the results of question-4 for those narratives.

\begin{figure} [H] 
    \centering
    \includegraphics[width=\linewidth]{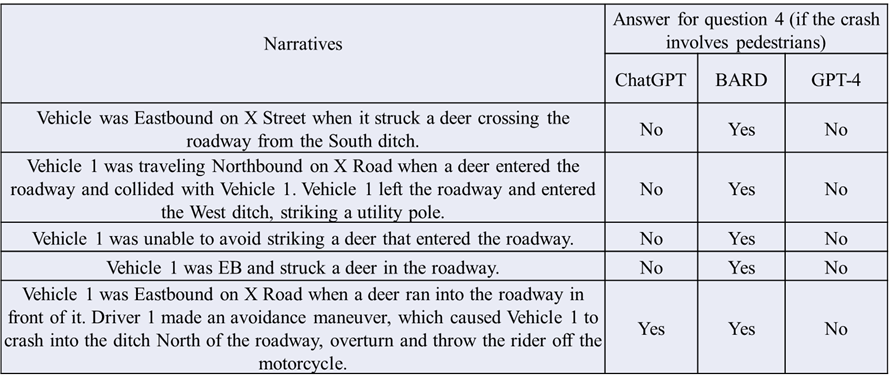}
    \caption{Narratives indicating deer-collision.
}
\end{figure}

Considering the answers for question-4, it appears that GPT-4 provides more reliable answer than the other two interfaces. 

\subsection{Query 5}

In question-5, the LLM interfaces were asked to create the sequence of events from the narratives. As we have collected the narrative data from two different states – Iowa and Kansas, creating a network of sequence of events for all 100 crash narratives will make the network diagram very complex and it will be harder to extract any useful information as the dynamics of these 100 crashes might be very different. Hence, for question-5, we have considered 15 crash narratives from Iowa to create network diagrams for the sequence of events collected from the LLM interfaces. 

Figure-4, 5 and 6 shows the network diagram of sequence of events for ChatGPT, BARD and GPT-4 respectively. There are some words which are abbreviated in these diagrams to show the network more clearly. The abbreviated parts are as follows- ‘PC’: Pre-crash event, ‘NC’: Non-collision Event, ‘C’: Collision with, ‘$C\_F O$’: Collision with Fixed Object, ‘ME’: Miscellaneous event. In the diagram, each node is placed at the center of the labelled text for that particular node. In figure-4, the network diagram for ChatGPT shows ‘Collision with vehicle in traffic’ is higher in degree centrality. And the most prominent edge is ‘Pre-crash event: Crossed centerline (undivided)’ to ‘Collision with vehicle in traffic’.

These type of frequent patterns from the network diagram will help the crash diagnostics of any particular location. In figure-5, there are several prominent nodes in the network diagram from BARD. ‘Collision with fixed object: Ditch’ and ‘Collision with vehicle in traffic’ are the most prominent nodes in BARD network diagram. It is also observed that the most prominent edge in the BARD diagram is the same as the ChatGPT diagram. In figure-6 for GPT-4, the most prominent node is ‘Collision with vehicle in traffic’ which is the same as ChatGPT. And the edge that stands out most in GPT-4 is also consistent with the other two LLM interfaces. Apart from these, there remain some major differences. For example, both ChatGPT and GPT-4 network diagram (in figure 4 and 6) have the node ‘Collison with Railway vehicle/train’ but while checking the narrative it was found that the crash occurred between two vehicles on a railroad crossing but it does not involve any railway vehicle. In BARD network diagram (Figure-5), the node ‘Pre-crash event: Loss of traction’ was one of the prominent nodes but it is not obvious from the narratives whether it is loss of traction or not. It is worth noting that, overall, BARD tends to use some events in the nodes extensively, giving more weight to those events, whereas with ChatGPT and GPT-4, the weight of the events is more distributed and varied.

\setlength{\intextsep}{2pt} 
\begin{figure} [H] 
    \centering
    \setlength{\intextsep}{0pt} 
    \includegraphics[width=\linewidth]{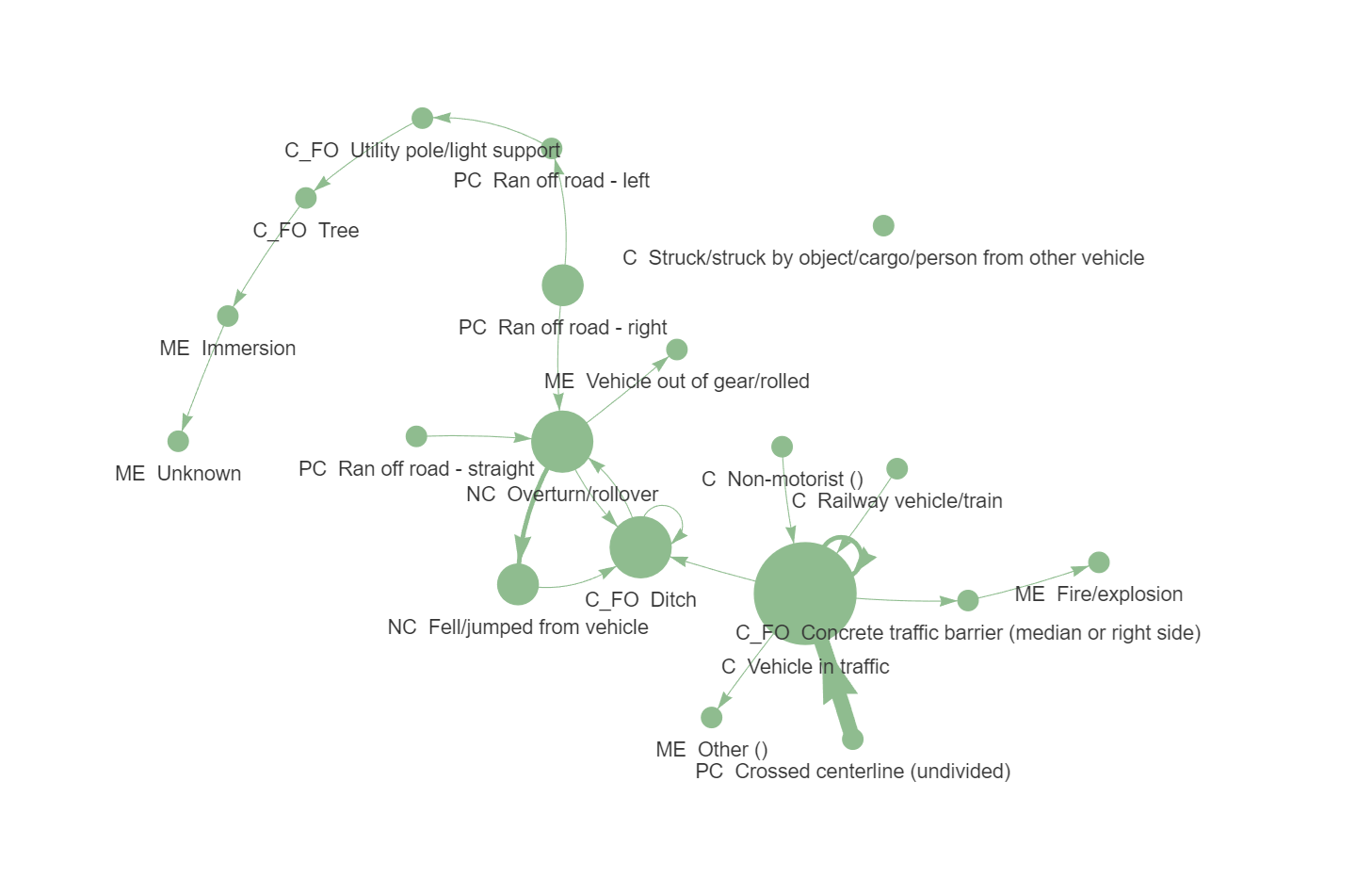}
    \caption{Network diagram for ChatGPT (Question 5- sequence of events), [PC: Pre-crash event, NC: Non-collision Event, C: Collision with, ‘$C\_F O$’: Collision with Fixed Object, ME: Miscellaneous event]
    \setlength{\intextsep}{5pt} 
}
\end{figure}

\setlength{\intextsep}{2pt} 

\begin{figure} [H] 
    \centering
    \setlength{\intextsep}{2pt} 
    \includegraphics[width=\linewidth]{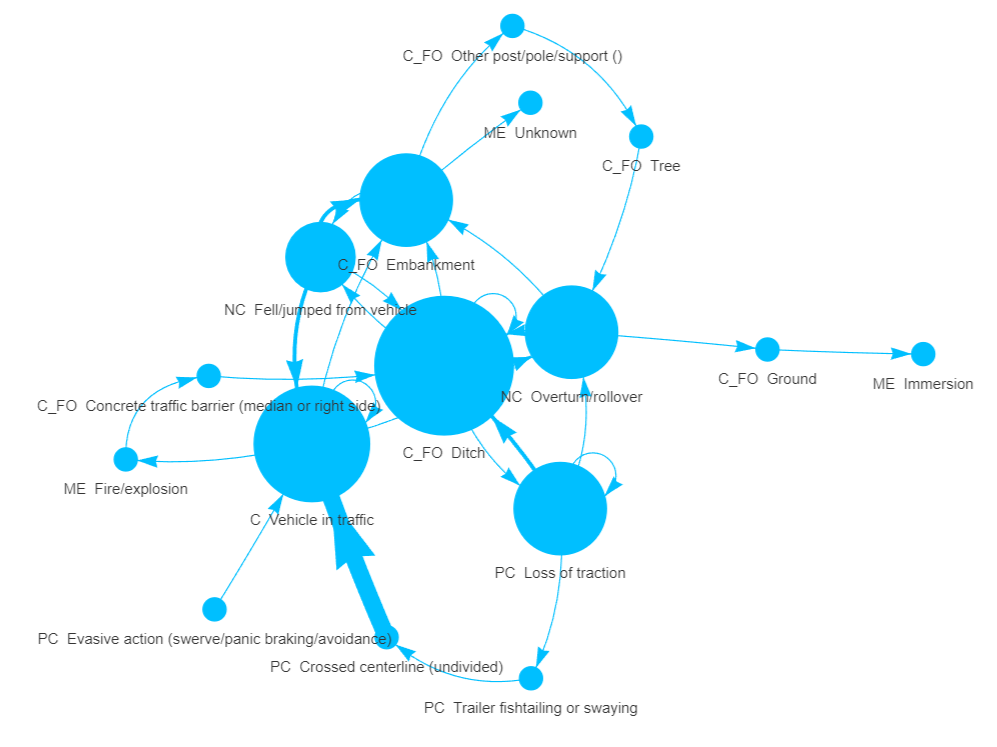}
    \caption{Network diagram for BARD (Question 5- sequence of events), [PC: Pre-crash event, NC: Non-collision Event, C: Collision with,‘$C\_F O$’: Collision with Fixed Object, ME: Miscellaneous event]
}
\end{figure}

\setlength{\intextsep}{2pt} 

\begin{figure} [H] 
    \centering
    \includegraphics[width=\linewidth]{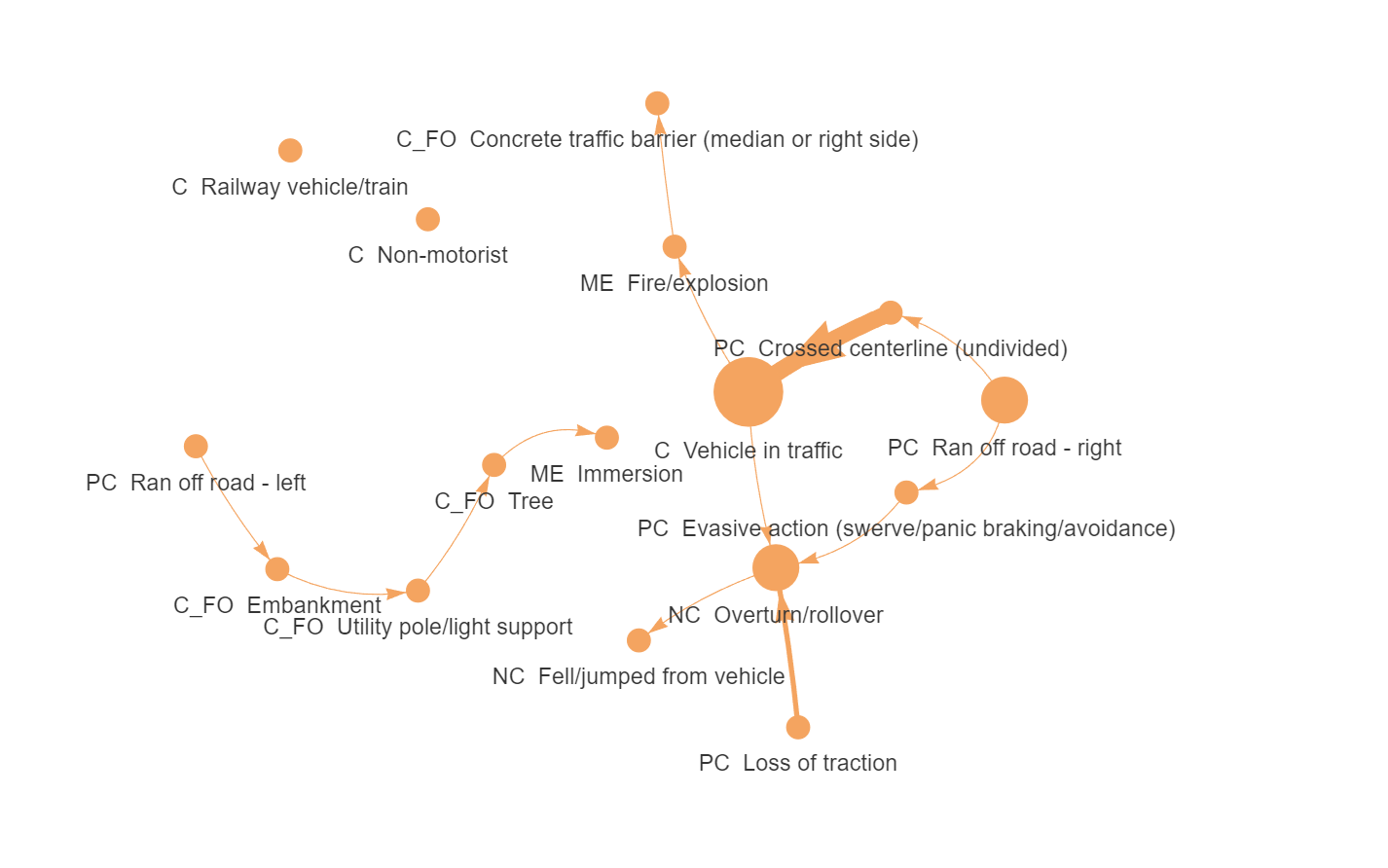}
    \caption{Network diagram for GPT-4 (Question 5- sequence of events), [PC: Pre-crash event, NC: Non-collision Event, C: Collision with, ‘$C\_F O$’: Collision with Fixed Object, ME: Miscellaneous event]
}
\end{figure}

To understand the most influential nodes in the network diagram, the centrality measures were calculated. Table-4 shows the top five nodes from each network diagram by considering in-degree, out-degree and betweenness centrality. Higher in-degree centrality means this is the most common outcome due to the other events. Although not identical, events like ‘Collision with Vehicle in traffic’, ‘Collision with Fixed Object: Ditch’, ‘Non-collision Event: Overturn/rollover’ were found to be higher in degree centrality in all three LLMs. So, the events surrounding these high in-degree centrality nodes should be examined to understand the crash at a particular location. In this study, the events with higher in-degree centrality are also found to have higher out-degree centrality. This means that these events are very frequent and influential in the crash events. Moreover, these events also have higher weights in betweenness centrality, thus cementing the proposition that these events are mostly controlling all other events in the crash sequence. From Table-4, it can be observed that for out-degree and betweenness centrality, ChatGPT has more similarity with GPT-4 for top serving nodes. For in-degree centrality, the similarity is higher among ChatGPT and BARD.

\begin{table}[H]
    \centering
    \caption{Centrality measures from the networks of sequence of events. [PC: Pre-crash event, NC: Non-collision Event, C: Collision with, C\_FO: Collision with Fixed Object, ME: Miscellaneous event]}
    \begin{tabularx}{\linewidth}{>{\centering\arraybackslash}p{2.5cm}X X X}
        \toprule
        \textbf{Centrality} & \textbf{ChatGPT} & \textbf{BARD} & \textbf{GPT-4} \\
        \midrule
        \multicolumn{4}{l}{\textbf{In-degree}} \\
        & C: Vehicle in traffic & C\_FO: Ditch & NC: Overturn/rollover \\
        & C\_FO: Ditch & C: Vehicle in traffic & PC: Crossed centerline (undivided) \\
        & NC: Overturn/rollover & C\_FO: Embankment & C: Vehicle in traffic \\
        & C\_FO: Concrete traffic barrier (median or right side) & NC: Overturn/rollover & ME: Fire/explosion \\
        & PC: Ran off road - left & NC: Fell/jumped from vehicle & C\_FO: Embankment \\
        \midrule
        \multicolumn{4}{l}{\textbf{Out-degree}} \\
        & C: Vehicle in traffic & C\_FO: Ditch & C: Vehicle in traffic \\
        & NC: Overturn/rollover & C: Vehicle in traffic & NC: Overturn/rollover \\
        & C\_FO: Ditch & C\_FO: Embankment & PC: Ran off road - right \\
        & PC: Ran off road - right & PC: Loss of traction & PC: Crossed centerline (undivided) \\
        & NC: Fell/jumped from vehicle & NC: Overturn/rollover & ME: Fire/explosion \\
        \midrule
        \multicolumn{4}{l}{\textbf{Betweenness}} \\
        & C: Vehicle in traffic & C\_FO: Ditch & C: Vehicle in traffic \\
        & NC: Overturn/rollover & NC: Overturn/rollover & NC: Overturn/rollover \\
        & C\_FO: Ditch & C: Vehicle in traffic & C\_FO: Utility pole/light support \\
        & C\_FO: Utility pole/light support & C\_FO: Embankment & PC: Crossed centerline (undivided) \\
        & C\_FO: Tree & NC: Fell/jumped from vehicle & ME: Fire/explosion \\
        \bottomrule
    \end{tabularx}
\end{table}

After exploring all these three interfaces, it is observed that the answers are not always consistent among these three interfaces which questions their credibility in identifying critical information. However, it was also found that when these three interfaces give the same answer, it is most likely to capture the actual information from the narratives. This study suggests usage of multiple models by leveraging their strengths to achieve more comprehensive and accurate analysis. In text analysis, a lot of the time, once the classification is done from any model, it is usually re-evaluated by humans to check for the correctness of the models. In this context, if the results are similar from multiple interfaces, it will require less human effort for quality control. 

This study has some notable limitations. Although the research goal was to explore the ability of LLMs to extract information from narratives, the answers were not validated with actual crash database. In future, it can be checked with other NLP models and ground truth value to evaluate the correctness of the answers. Another major limitation is- these interfaces are not consistent in providing results so using them will not ensure reproducibility of the results all the time. Hence, caution must be practiced while using these LLMs for any safety related issues. In this study, we have used the LLM interfaces directly. But, in any future research the process can be automated by using the API of these LLM interfaces. 

\section{CONCLUSION}

We have considered 100 crash narratives to explore the capabilities of the LLMs for extracting information related to crash narratives. This study has three major contributions: 1) It shows a process regarding using the LLM interfaces to extract concise information from crash narratives by providing clear guidelines in the prompt, 2) It discusses the effectiveness of different LLM interfaces addressing a range of inquiries, including some complex ones, and 3) It shows the similarities, dissimilarities, and limitations of different LLM’s responses related to crash narratives. This current study also provides the rationale behind each question asked to the LLMs. It is needful to know how the state and local agencies use crash narratives to extract information and the similar queries should be tried in text mining. For example, while the narratives usually hold information related to fatalities on spot, determining the other injury type (major/minor/possible) from these narratives is typically challenging. So, employing a model to classify all injury types is not a viable option. This study has observed that the similarities between responses among these interfaces differed with each question. The overall similarity in determining at-fault was 70\% and 'Vehicle-1' was considered at fault in most of the cases by all three LLMs. Work-zone occurrence and pedestrian involvement related responses were 96\% and 89\% similar, respectively, among the three LLM interfaces. Indication of ‘construction zone’ was picked up by all three LLMs successfully while detecting work-zone related crashes. In some cases, indication of barrier wall, crash barrel, patrol vehicle with emergency lights on and driver removing debris from the roadway was also detected as work-zone crashes by some LLMs. While asked to detect pedestrian, BARD has classified crash narratives indicating ‘deer’ as pedestrian crashes. GPT-4 was found to be most reliable in pedestrian crash detection. The question determining the manner of collision shows 35\% similarity only as it required more complex responses from the LLMs. For question 5, sequence of events from crash narratives, network diagrams along with centrality measures were developed and the observed similarities were not same in all instances. Network analysis from all three LLM interfaces resulted in harmful events like ‘Collision with Vehicle in traffic’, ‘Collision with Fixed Object: Ditch’, and ‘Non-collision Event: Overturn/rollover’ to be higher in degree centrality in the event of a crash. It is notable that, the similarities were higher while answering simple binary questions such as pedestrian involvement or work-zone crash. In contrast, while dealing with more complex information like manner of collision or sequence of events, the similarity was significantly lower. For any future studies, it is suggested to combine results from multiple models to extract viable information from crash narratives.

\bibliographystyle{unsrt}  
\bibliography{references}  


\end{document}